\documentclass{article}

\usepackage[square,numbers]{natbib}
\usepackage[preprint]{neurips_2025}

\usepackage{subfig}
\usepackage[utf8]{inputenc} 
\usepackage[T1]{fontenc}    
\usepackage{hyperref}       
\usepackage{url}            
\usepackage{booktabs}       
\usepackage{amsfonts}       
\usepackage{nicefrac}       
\usepackage{microtype}      
\usepackage{xcolor}         
\usepackage{adjustbox}

\usepackage{hyperref}
\usepackage{tabularx}
\usepackage{wrapfig}
\usepackage{url}
\usepackage{multirow}
\usepackage{graphicx}
\newcommand{\cut}[1]{}

\usepackage{makecell}

\title{Infinity Instruct: Scaling Instruction Selection and Synthesis to Enhance Language Models}

\author{%
  Jijie Li$^*$, \space Li Du\thanks{Core contributors with equal contributions.}, \space Hanyu Zhao, \space Bo-wen Zhang, \space Liangdong Wang, \\
  \textbf{Boyan Gao}$^\xi$,\space \textbf{Guang Liu}\thanks{Project Lead, the corresponding author, contact liuguang@baai.ac.cn}, \space \textbf{Yonghua Lin} \\
  Beijing Academy of Artificial Intelligence, $^\xi$University of Oxford \\
}

\begin{document}

\maketitle
\begin{abstract}

Large Language Models (LLMs) demonstrate strong performance in real-world applications, yet existing open-source instruction datasets often concentrate on narrow domains, such as mathematics or coding, limiting generalization and widening the gap with proprietary models. To bridge this gap, we introduce Infinity-Instruct, a high-quality instruction dataset designed to enhance both foundational and chat capabilities of LLMs through a two-phase pipeline. In Phase 1, we curate 7.4M high-quality foundational instructions (InfInstruct-F-7.4M) from over 100M samples using hybrid data selection techniques. In Phase 2, we synthesize 1.5M high-quality chat instructions (InfInstruct-G-1.5M) through a two-stage process involving instruction selection, evolution, and diagnostic filtering. We empirically evaluate Infinity-Instruct by fine-tuning several open-source models, including Mistral, LLaMA, Qwen, and Yi, and observe substantial performance gains across both foundational and instruction following benchmarks, consistently surpassing official instruction-tuned counterparts. Notably, InfInstruct-LLaMA3.1-70B outperforms GPT-4-0314 by 8.6\% on instruction following tasks while achieving comparable foundational performance. These results underscore the synergy between foundational and chat training and offer new insights into holistic LLM development. Our dataset\footnote{https://huggingface.co/datasets/BAAI/Infinity-Instruct} and codes\footnote{https://gitee.com/li-touch/infinity-instruct} have been publicly released.
\end{abstract}

\section{Introduction}

The advent of Large Language Models (LLMs) marks a significant milestone in artificial intelligence, with instruction fine-tuning emerging as a crucial technique to unlock their full potential in natural language processing by enhancing their ability to follow complex prompts~\cite{zhang2024aquila2,ouyang2022training,yang2024qwen2}. As these models become increasingly central to real-world applications, the quality and diversity of their training data, especially instruction data, have become key determinants of their performance~\cite{OpenHermes25,zheng2023lmsyschat1m,xu2023wizardlm,li2023quantity}.

While fine-tuning LLMs on instruction data improves task-specific performance, naively adapting pretrained models to downstream instructions without sufficient regularization can lead to catastrophic forgetting of fundamental linguistic and reasoning capabilities~\cite{ding2023enhancing}. This presents a delicate trade-off between enhancing task alignment and preserving core generalization abilities.

Constructing comprehensive, high-quality instruction datasets from scratch for fine-tuning, especially at the scale required by modern LLMs, is prohibitively expensive in terms of both human labor and computational resources. A tempting alternative is to combine existing open-source datasets. However, this approach is fraught with challenges: the quality of these datasets is not guaranteed, random combinations may degrade performance, and effective mixing often requires expert knowledge and time-consuming manual tuning of dataset composition ratios.

To address these issues, we propose Infinity-Instruct, a principled and scalable pipeline for systematically constructing instruction datasets by solving key challenges in instruction selection, labeling, and synthesis. Rather than relying on heuristic combinations, we introduce a labeling-driven strategy to select and synthesize data from a large instruction pool, yielding datasets that are both diverse and high-quality. We release two curated datasets:
\begin{itemize}  
 \item InfInstruct-F-7.4M, a foundational instruction dataset consisting of 7.4M instructions selected using a blend of filtering, clustering, and coverage-based strategies.

 \item InfInstruct-G-1.5M, a general-purpose conversational instruction dataset built on a two-layer instruction labeling system. We first cluster and label instructions using an open-source model, select 1.2M high-quality seeds, and then expand them via iterative synthesis and diagnosis to create a richer set of 1.5M conversational instructions.
\end{itemize} 
These datasets undergo rigorous de-duplication and contamination filtering to ensure clean generalization. To evaluate the effectiveness of Infinity-Instruct, we fine-tune several popular open-source LLMs and produce the InfInstruct model family, including InfInstruct-Mistral-7B, InfInstruct-Llama3.1-8B/70B, InfInstruct-Qwen-2-7B, and InfInstruct-Yi-1.5-9B. These models consistently outperform their official instruction-tuned counterparts on both foundational and conversational benchmarks. Notably, InfInstruct-Llama3.1-70B surpasses GPT-4-0314 by 8.6\% in conversational ability and matches its performance in foundational tasks, while InfInstruct-Llama3.1-8B achieves gains of 4.4\% and 7.4\% on foundational and conversational benchmarks, respectively.

Our findings also reveal a positive correlation between foundational and conversational capabilities, emphasizing the importance of balanced dataset design. Infinity-Instruct offers a scalable solution to instruction dataset construction, bridging the gap between open-source and proprietary LLM performance. Our main contributions are summarized as follows:  
\begin{itemize}  
    \item \textbf{Introduction of a unified dataset construction pipeline}: We design a novel pipeline that systematically curates datasets and synthesizes data to address both foundational and conversational tasks, ensuring high quality and diversity.  

    \item \textbf{Development of a high-quality dataset, Infinity Instruct}:  
    \begin{itemize}  
        \item \textbf{InfInstruct-F-7.4M}: A foundational dataset of 7.4M instructions selected from over 100M samples using robust data selection strategies.  
        \item \textbf{InfInstruct-G-1.5M}: A conversational dataset synthesized through a two-layer labeling system and iterative refinement, ensuring diversity and quality in conversational scenarios.  
    \end{itemize}  

    \item \textbf{Ensuring data quality and generalization}: We apply de-duplication and contamination detection techniques to guarantee robustness and applicability across domains such as math, code, and knowledge Q\&A.  

    \item \textbf{Demonstration of superior model performance}: Fine-tuning multiple open-source models (e.g., Mistral, Llama, Qwen, and Yi) on Infinity-Instruct achieves state-of-the-art results on foundational and conversational benchmarks. Notably, \textbf{InfInstruct-Llama3.1-70B} surpasses GPT-4-0314 by 8.6\% in conversational ability and achieves near parity in foundational tasks.  

\end{itemize}


\section{Methodology}
\begin{figure*}[htbp]
    \centering
    \includegraphics[width=0.9\linewidth]{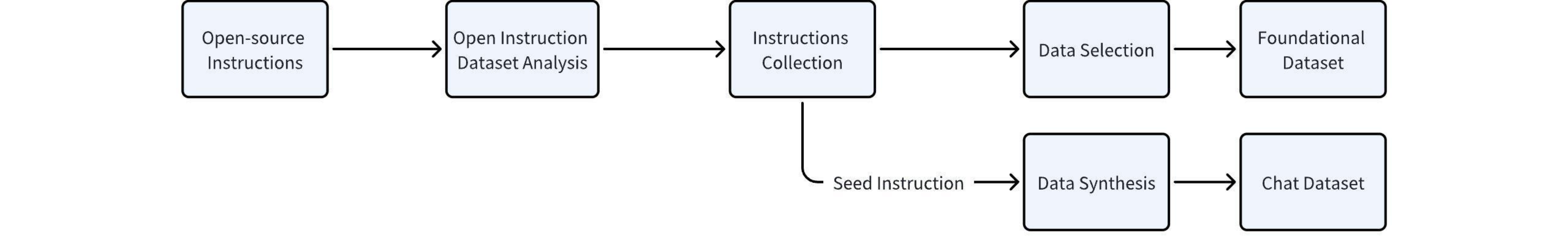}
    \caption{The overall structure for building the Infinity-Instruct dataset.}
    \label{fig:framework}
\end{figure*}
The Infinity-Instruct pipeline, as depicted in Figure~\ref{fig:framework}, we first analyze the shortcomings of the open source instruction dataset based on fine-tuning experiments. Then we over 100 million open-source instructions from diverse domains such as math, code, and knowledge Q\&A. The pipeline then follows a two-stage process. First, a rigorous data selection module is applied to filter and curate high-quality foundational instructions, resulting in the foundational dataset (InfInstruct-F-7.4M). Second, a subset of seed instructions is identified and subjected to a data synthesis stage. This stage employs iterative instruction labeling, evolution, and diagnostic processes to generate the conversational dataset (InfInstruct-G-1.5M), ensuring robustness and diversity in conversational capabilities. This structured approach ensures that both foundational and conversational datasets are optimized for enhancing LLM performance comprehensively.

\subsection{Open instruction dataset analysis}
\label{open source experiments}

There are consistent differences between the performance of open-source and closed-source models in real scenarios. We believe that an important gap stems from the instruction dataset. To confirm the inadequacy of the open-source instruction dataset, we selected Mistral-7B to be fine-tuned on several recent open-source datasets. We used the overall capability of GPT-3.5 as a comparison to find the overall capability shortfall of the fine-tuned Mistral-7B. We first collected a collection of popular language modeling benchmarks.

The results, summarized in Table \ref{Commonsense ablation}, reveal that OpenHermes consistently delivered the best performance among open-source datasets. However, a significant gap remains when compared to closed-source models such as GPT-3.5 and GPT-4, particularly in areas such as data diversity, code comprehension, knowledge-based quizzes, dialog generation, and other advanced capabilities.

\subsection{Instruction Collection}

\begin{figure}[htbp]
    \centering
    \includegraphics[width=0.9\linewidth]{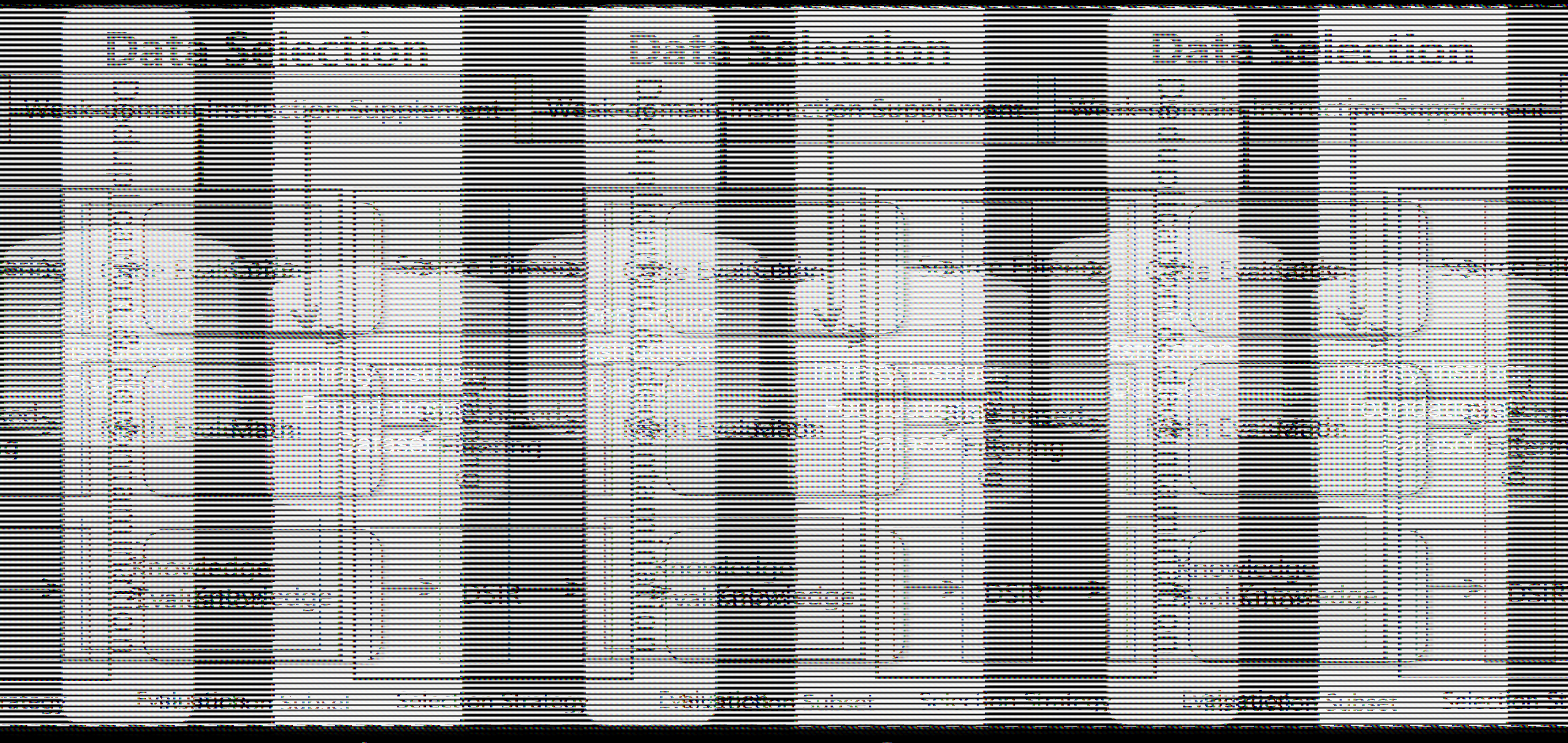}
   
    \caption{Overall Pipeline of Data Selection Pipeline}
    \label{fig:data-select}
\end{figure}

Based on the open instruction dataset analysis, we first collected some of the most current and up-to-date instruction datasets in the fields of math, code, commonsense question-answering, and conversational to construct the data pool. This data pool contains more than 100 million instructions. The statistics of the data pool are shown in Table \ref{domain statistics}.


We selected some recent datasets from the data pool and applied Mistral-7B-v0.1 to fine-tune these datasets. The result is shown in Table \ref{Commonsense ablation}. We found that OpenHermers can achieve the overall best results. However, there is still a gap compared to GPT-3.5 or GPT-4, which motivates us to further improve the quality and diversity of the existing instruction datasets through data selection and data synthesis. 

\begin{table}[tb]
\centering
\small
\caption{Source data and used data statistics of Infinity-Instruct}
\begin{tabular}{lcc}
\hline
Domain             & Total Collection (M) & Used (M) \\ \hline
Code               & 7.1                  & 1.5                            \\
Math               & 11.8                 & 1.4                            \\
Knowledge          & 88.5                 & 3.3                            \\
Instruction Follow & 9.0                  & 2.8                            \\
Total              & 116.4                & 9.0                            \\ \hline
\end{tabular}
\label{domain statistics}
\end{table}



\subsection{Data Selection}
\label{data-select}

In this chapter, we will introduce the data selection module used to build the foundational dataset of Infinity-Instruct. The pipeline of this module is demonstrated in Figure \ref{fig:data-select}. 

\textbf{Selection strategy}. We introduce three selection strategies such as source filtering, rule-based filtering, and DSIR~\cite{xie2023dataselectionlanguagemodels}. From a task perspective, we choose the appropriate filtering rule according to the features of the tasks such as code and math. The details of the selection strategy used for each task are described as follows:

\begin{itemize}
    \item \textbf{Knowledge}. We found that the collected general knowledge datasets were of varying quality due to different data cleaning rules. To enhance the model’s capacity for commonsense understanding, we introduce the Flan 2022 dataset~\cite{longpre2023flan}, which consolidates all publicly available academic datasets for instruction tuning at the time. This dataset integrates hundreds of high-quality templates, diverse formatting patterns, and extensive data augmentation, including a mix of zero-shot, few-shot, and chain-of-thought prompt formats. We specifically curated the dataset by filtering out samples from sources with relatively low knowledge content, such as sentiment classification datasets (e.g., SST-2, IMDb movie reviews). Additionally, for synthetic datasets and augmented data samples (e.g., question-answering and question-generation samples generated from the same seed data), we implemented a deduplication strategy to reduce the proportion of augmented samples.

    \item \textbf{Math}. We refer to the DSIR~\cite{xie2023dataselectionlanguagemodels}, which aims at selecting a subset of a large raw unlabeled dataset to match a desired target distribution given unlabeled target samples. To improve the model's mathematical abilities, we will use the prompts from GSM8K and MATH training samples as target distributions to guide subset selection from the pool of mathematical datasets. In addition to selecting from existing datasets, to enhance the model's sensitivity to numerical variations, we synthesize data based on current datasets. This involves generating corresponding chain-of-thought (CoT) and program-of-thought (PoT) reasoning processes for mathematical problems, along with using data augmentation strategies to expand the dataset. Details can be found in \cite{zhang2024infinitymath}. In the end, we constructed about 1.4M math instruction datasets.
    
    \item \textbf{Code}. In line with our methodology for enhancing mathematical understanding, we apply the DSIR technique to curate data from an extensive collection of open-source instruction datasets in the coding domain. To ensure that selected data aligns closely with the target task characteristics, we base the importance resampling process on prompt distributions obtained from HumanEval samples. This approach allows us to prioritize samples that better represent the types of coding challenges and reasoning processes seen in HumanEval, ultimately refining the model’s ability to generalize and perform across diverse coding tasks.
\end{itemize}

\textbf{Evaluation and Weak-domain Instruction Supplement}. To optimize data utilization, we adopt a gradual approach by incrementally increasing the data volume for each task. As described in section \ref{open source experiments}, we evaluate the saturation level of the current dataset for each task by fine-tuning experiments in Mistral-7B. When a performance gap is observed between the fine-tuned Mistral-7B on the current dataset version and a baseline model (e.g., GPT-3.5) for a specific task, we will relax the data selection criteria to incorporate additional weak-domain data, thereby enhancing the model's performance in underrepresented areas.

Finally, by merging the foundational instructions of each task, we obtained 6.2M generalized foundational instruction datasets. Besides, to ensure smooth switching between foundation training and instruction-following training, similar to the replay strategy \citep{zhang2024map}, we added 1.2M seed instructions (as mentioned in \ref{seed instruction}) to make up the final version foundational instructions dataset, InfInstruct-F-7.4M. We find that this strategy can guide the model to achieve the optimal foundational ability and the instruction-follow performance.

\begin{table}[tb]
\centering
\small
\caption{Downstream task performance. We illustrate the performance of the task specific models developed from Mistral-7B by fine-tuning on open-source candidate datasets, including GSM-8K, MATH, HumanEval, BMPP, MMLU and C-EVAL. }
\begin{tabular}{lcccccc}
\toprule
\multicolumn{1}{c}{\multirow{2}{*}{}} & \multicolumn{2}{c}{Math} & \multicolumn{2}{c}{Code} & \multicolumn{2}{c}{Knowledge} \\ \cline{2-7} 
\multicolumn{1}{c}{}                  & GSM-8K       & MATH      & HumanEval     & MBPP     & MMLU           & C-EVAL          \\ \hline
GPT-3.5 & 57.1    & \textbf{28.0}  & \textbf{48.1}  & \textbf{68.2} & \textbf{70.0} & \textbf{54.4} \\
Mistral-7B                            & 48.1         & 11.8      & 14.0          & 38.0     & 56.5           & 34.6           \\
w/ OpenHermers                       & \textbf{73.0}         & 18.2      & 41.5          & 41.8     & 61.7           & 43.0           \\
w/ Mammoth-v2                         & 50.0         & 17.6      & 27.4          & 33.0     & 57.9           & 40.4           \\
w/ Bagel                              & 58.9         & 18.3      & 35.4          & 39.2     & 54.9           & 37.5           \\ 
\bottomrule
\end{tabular}
\label{Commonsense ablation}
\end{table}

\begin{figure}[ht]
    \centering
    \includegraphics[width=0.9\linewidth]{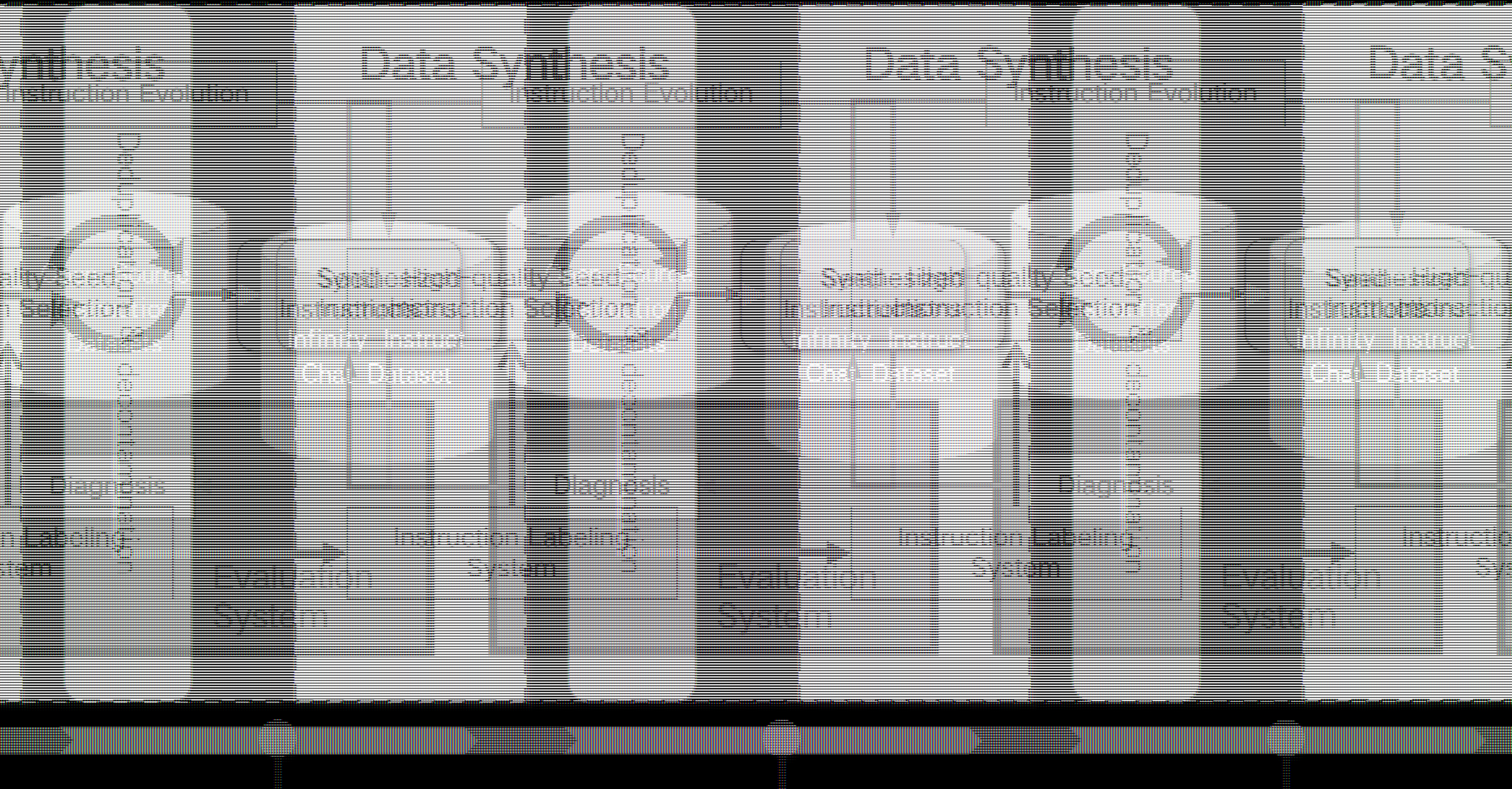}
    \caption{Overall Pipeline of Data Synthesis Pipeline}
    \label{fig:data-synthesis}
\end{figure}

\subsection{Data Synthesis}
\label{synthesis}

The existent finding diverse and difficult dialog instructions helps to improve the model's ability to converse in real-life scenarios. In this chapter, we design an iterative instruction evolution pipeline. As shown in Figure \ref{fig:data-synthesis}, this pipeline contains four steps of instruction labeling system construction, high-quality seed instruction construction, instruction evolution, and model weakness detection. We started with 9M open source instructions and eventually synthesized a dataset of about 1.5M evolutionary instructions, namely InfInstruct-G-1.5M.

\textbf{Instruction Labeling System}. A high-quality labeling system facilitates the synthesis of diverse data and improves the generalization of the model across different conversation scenarios. So we use an open-source language model (i.e., Qwen1.5-72B) to label each instruction at the second level, and then regularize the second-level labels through embedding clustering and manual tuning. Eventually we will further generalize the first level labels using the generalization ability of the language model. The final labeling system we constructed contains 26 first-level labels and more than fifteen thousand second-level labels. Details about the labeling system can be found in ~\cite{zhao2024iidoptimizinginstructionlearning}.




\label{seed instruction}
\textbf{High-quality Seed Instruction Selection}. We apply difficulty and diversity as criteria and select 1.2M high quality dialog instructions from 9M dialog instructions. We uniformly used Qwen 1.5-7B for the evaluation of instruction difficulty. Specifically, the filtering guidelines are as follows: (1) \textbf{Long-tail data (Diversity)}. Based on the labeling system, we retained all the instructions in the ability labels with a frequency of occurrence between 20 and 200, and extracted 1/3 of the instructions for each ability label with a frequency between 200 and 500. These instructions will not enter the subsequent filtering process. (2) \textbf{Multidimensional capabilities. (Diversity)}. As discussed in~\cite{zhao2024iidoptimizinginstructionlearning}, we consider the instruction involved multiple capabilities models a difficult task. Based on the labeling system, we prioritized the retention of instructions involving multiple capability labels. (3) \textbf{High language modeling loss (Difficulty).} We compute the loss for the answer portion of a candidate instruction. A higher loss reflects the model's lack of familiarity with the capabilities involved in the instruction, as well as the more difficult it is to provide a correct response to the instruction. (4) \textbf{High convergence loss (Difficulty)}. As discussed in~\cite{li2023quantity}, learning on certain instructions may easily lead to overfitting, causing the model to develop harmful biases that affect the generalizability. To avoid this phenomenon, we removed samples that had been modeled with a large difference in loss before and after fine-tuning.

\textbf{Instruction Evolution}. We further expand the difficulty of the seed data by multi-step rewriting through the evolve-instruct instruction evolution algorithm proposed by ~\citep{xu2023wizardlm}. For each seed instruction, we applied Wizard's four evolutionary strategies to rewrite it, and 
 ask the rewriting model to determine whether the rewritten instruction was semantically identical to the one before rewriting or whether it introduced harmful information.

\textbf{Diagnosis}. Finally, we attempt to analysis the weak ability of open-source models based on the instruction labeling system and supplemented with synthesized instructions for theses ability types. Specifically, we extracted evolved instructions from each ability type and applied GPT-4 to evaluate the quality of responses from multiple open-source models (including Mistral-7B, Llama3-8B). Instructions that result in poor responses from either model are added to the next round of instruction evolution. This feedback-based strategy guarantees that our instruction dataset scaling more efficiently.

\subsection{Deduplication and Decontamination}
To avoid any duplication, we decontaminate and de-duplicate the synthetic data. We use the BGE~\cite{bge_embedding} model to vectorize the instructions and calculate the cosine similarity between the instructions in the 
Infinity-Instruct and the open-source benchmark. We manually confirmed a threshold of 0.3 for filtering duplicates and contamination.

\section{Experiments}

In this section, we empirically evaluate the effect of Infinity-Instruct with existing large-scale instruction datasets and verify the stability of Infinity-Instruct over different models.
\begin{table}[tb]
\centering
\caption{Infinity Instruct dataset's enhancement of open source model conversational capabilities.}
\begin{tabular}{lcccc}
\hline
\textbf{Dataset}           & \textbf{AlpacaEval 2.0}   & \textbf{Arena-Hard}  & \textbf{MT-Bench}  & \textbf{Overall*}    \\ \hline
\multicolumn{4}{l}{\textit{\textbf{Official closed-source models}}}                                & \multicolumn{1}{l}{} \\ \hline
GPT-3.5-0301               & 18.1                      & 29.4                 & 7.9                & 42.2                 \\
GPT-4-0314                 & 35.3                      & 50.0                 & \textbf{9.0}       & 58.4                 \\ \hline
\multicolumn{4}{l}{\textit{\textbf{Official open-source models}}}                                  & \multicolumn{1}{l}{} \\ \hline
Mistral-7B-Instruct-v0.2   & 17.1                      & 16.2                 & 7.6                & 36.4                 \\
Qwen2-7B-Instruct          & 20.9                      & 19.6                 & 8.4                & 41.5                 \\
Yi-1.5-9B-conversational             & 20.5                      & 16.3                 & 8.0                & 38.9                 \\
Llama3.1-8B-Instruct       & 20.9                      & 20.6                 & 8.1                & 40.8                 \\
Llama3.1-70B-Instruct      & 38.1                      & 55.7                 & 8.6                & 59.9                 \\ \hline
\multicolumn{4}{l}{\textit{\textbf{Official open-source models finetuned with Infinity-Instruct}}} & \multicolumn{1}{l}{} \\ \hline
InfInstruct-Mistral-7B     & 40.0                      & 26.9                 & 8.1                & 49.3 (+12.9)         \\
InfInstruct-Qwen2-7B       & 21.9                      & 22.3                 & 8.3                & 42.4 (+0.9)          \\
InfInstruct-Yi-1.5-9B      & 20.5                      & 17.4                 & 8.0                & 39.3 (+0.4)          \\
InfInstruct-Llama3.1-8B    & 33.9                      & 33.7                 & 8.2                & 49.9 (+9.1)          \\
InfInstruct-Llama3.1-70B   & \textbf{46.1}             & \textbf{66.0}        & 8.9                & \textbf{67.0 (+7.1)} \\ \hline
\end{tabular}
\label{chateval}
\end{table}

\subsection{Comparison with official aligned models}
In this section, we finetune multiple open-source pre-trained models on Infinity-Instruct to verify the generalizability of Infinity Instruct. As shown in Table \ref{chateval}, we can see that after fine-tuning, Qwen2, Yi-1.5, Llama3.1 and mistral-v0.1 have all surpassed the conversational performance of the official instruct models. In addition, InfInstruct-Llama3.1-70B (46.1) even exceeds GPT4-0314 (35.3), which truly realizes the GPT-4 level of conversational ability. It is important to note that most officially released models have already undergone alignment to human preferences, which may further enhance the model's conversational capabilities. Besides, in Table \ref{foundeval}, we can find Infinity-Instruct helps to reproduce the foundational ability performance of the official conversational model. The fine-tuned versions of Llama3.1-8B and Mistral-7B-v0.1 on Infinity-Instruct comprehensively outperform the official versions.InfInstruct-Llama3.1-70B even outperforms GPT4-0314 in terms of code ability evaluation.


\subsection{Comparison with open source instruction datasets}     



\begin{table}[t]
\centering
\small
\caption{Infinity Instruct dataset's enhancement of open source model foundational capabilities.}
\adjustbox{max width=1.0\linewidth}{
\begin{tabular}{lccccccc}
\hline
\multirow{2}{*}{\textbf{Dataset}} & \multicolumn{2}{c}{\textbf{Math}} & \multicolumn{2}{c}{\textbf{Code}}  & \multicolumn{2}{c}{\textbf{Knowledge}} & \multirow{2}{*}{\textbf{Overall}} \\ \cline{2-7}
                                  & \textbf{MATH}  & \textbf{GSM-8K}  & \textbf{HumanEval} & \textbf{MBPP} & \textbf{MMLU}     & \textbf{C-EVAL}    &                                   \\ \hline
\multicolumn{7}{l}{\textit{\textbf{Official closed-source models}}}                                                                                  & \multicolumn{1}{l}{}              \\ \hline
GPT-3.5-0301                      & 28.0           & 57.1             & 48.1               & 60.2          & 70.0              & 54.4               & 53.0                              \\
GPT-4-0314                        & \textbf{45.8}  & \textbf{92.0}    & 67.0               & 61.8          & \textbf{86.4}     & 68.7               & \textbf{70.3}                     \\ \hline
\multicolumn{7}{l}{\textit{\textbf{Official instruct models}}}                                                                                    & \multicolumn{1}{l}{}              \\ \hline
Mistral-7B-Instruct-v0.2          & 9.5            & 45.9             & 32.9               & 3.8           & 59.6              & 42.6               & 32.4                              \\
Qwen2-7B-Instruct                 & 15.4           & 80.4             & 66.5               & 51.0          & 71.1              & \textbf{81.7}      & 61.0                              \\
Yi-1.5-9B-conversational                    & 17.8           & 78.7             & 65.3               & 46.5          & 62.1              & 62.7               & 55.5                              \\
Llama3.1-8B-Instruct              & 15.6           & 55.2             & 49.7               & 47.2          & 66.1              & 42.3               & 46.0                              \\
Llama3.1-70B-Instruct             & 43.3           & 91.7             & 67.7               & 63.2          & 82.2              & 62.3               & 68.4                              \\ \hline
\multicolumn{7}{l}{\textit{\textbf{Official foundation models finetuned with Infinity-Instruct}}}                                                   & \multicolumn{1}{l}{}              \\ \hline
InfInstruct-Mistral-7B            & 26.0           & 78.6             & 59.8               & 52.0          & 65.0              & 51.1               & 55.4 (+23.0)                      \\
InfInstruct-Qwen2-7B              & 15.2           & 83.2             & 66.5               & 54.4          & 70.4              & 81.5               & 61.9 (+0.9)                       \\
InfInstruct-Yi-1.5-9B             & 17.6           & 76.2             & 70.2               & 48.8          & 64.5              & 63.2               & 56.8 (+1.3)                       \\
InfInstruct-Llama3.1-8B           & 28.1           & 70.2             & 53.7               & 50.4          & 66.3              & 51.0               & 53.3 (+7.3)                       \\
InfInstruct-Llama3.1-70B          & 42.1           & 88.0             & \textbf{72.0}      & \textbf{70.3} & 79.0              & 63.1               & 69.1 (+0.7)                       \\ \hline
\end{tabular}
}
\label{foundeval}
\end{table}

In this section, we compare Infinity-Instruct as well as popular open-source datasets to open-source models in terms of their conversational, foundational ability to improve. 

\textbf{Conversational Ability Evaluation}. As shown in Table \ref{main-result}, we can see on the conversational ability evaluation, MAGPIE-Pro is by far the most effective dataset for improving the model's conversational competence, and the Llama 3.1-8B fine-tuned on top of it even outperforms the official version. However, Infinity-Instruct improves Llama3.1-8B more than MAGPIE-Pro on all three conversational ability review sets. 

\textbf{Foundational Ability Evaluation}. In terms of the foundational ability evaluation, as shown in Table \ref{main-result-2}, we can see that most of the datasets can improve the foundational capability of the base model. And fine-tuning on OpenHermes2.5, MAGPIE-Pro can further improve Llama3.1-8B to close to or even exceed the Llama3.1-8B-Instruct on some of the evaluation set. However, fine-tuning on Infinity-Instruct allows the Llama3.1-8B to surpass the Llama3.1-8B-Instruct's foundational capabilities across the board. It should thanks to the diversity and quality of the instructions of Infinity Instruct. 

\begin{table}[t]
\centering
\small
\caption{Infinity Instruct compares the effectiveness of open-source datasets in improving the conversational capabilities of pre-trained models. We tested the foundational and conversational abilities separately. The best scores are bolded. * To calculate the overall average score, we mapped MT-Bench scores to 0-100.}
\begin{tabular}{lcccc}
\toprule
\textbf{Dataset}                                                & \textbf{AlpacaEval} & \textbf{Arena-Hard} & \textbf{MT-Bench} & \textbf{Overall*} \\ 
\midrule
Llama3.1-8B-instruct                                            & 20.9                & 18.2                & 8.0               & 39.7             \\
WildChat~\cite{zhao2024wildchat}          & 11.4                & 6.7                 & 7.4               & 30.4             \\
Lmsys-conversational-1m~\cite{zheng2023lmsyschat1m} & 4.2                 & 2.2                 & 5.2               & 19.5             \\
Evol-Instruct~\cite{xu2023wizardlm}       & 2.0                 & 3.7                 & 6.6               & 23.9             \\
UltraChat~\cite{ding2023enhancing}        & 7.6                 & 3.5                 & 6.7               & 26.0             \\
MAGPIE-Pro~\cite{xu2024magpie}            & 28.0                & 22.6                & 8.2               & 44.2             \\
OpenHermes-2.5\cite{OpenHermes25}              & 13.1                & 14.2                & 7.5               & 34.1             \\
Infinity-Instruct                                               & \textbf{34.0}       & \textbf{38.9}       & \textbf{8.2}      & \textbf{51.6}    \\ \hline
\end{tabular}
\label{main-result}
\end{table}

\label{analysis}

\cut{
\begin{figure}[t]
    \centering
    \includegraphics[width=0.7\linewidth]{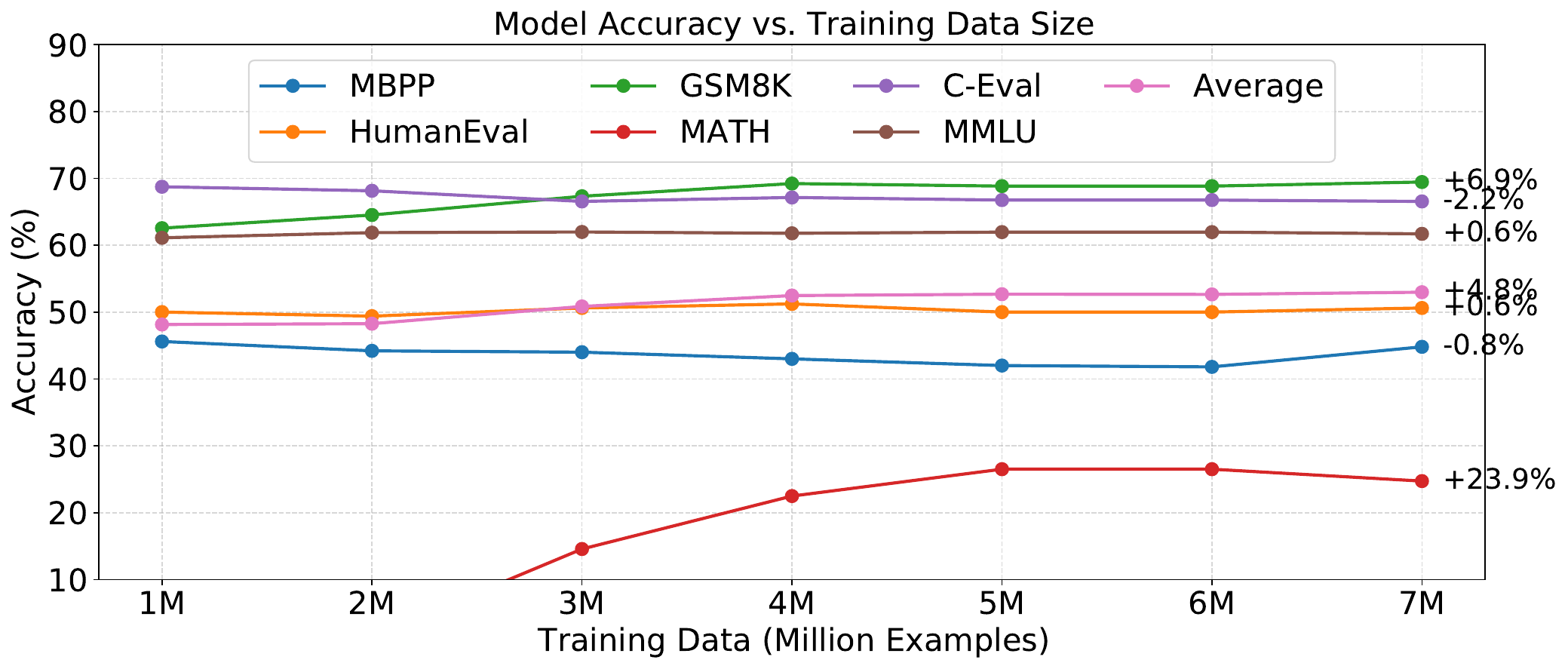}
    \caption{Performance dynamic against training dataset consumption. We record the model performance on a variety of benchmarks while training to demonstrate the benefit of Infinity Instruct.}
    \label{fig:task_improvement}
\label{tsne}
\end{figure}
}

\subsection{Scaling curves on foundational and conversational dataset} 
\begin{figure*}[t]
\centering
\subfloat[Foundational Tasks]{\includegraphics[width=0.53\linewidth]{infinity_instruct/dataset_accuracy_comparsion.pdf}}
\subfloat[Conversational Tasks]{\includegraphics[width=0.35\linewidth]{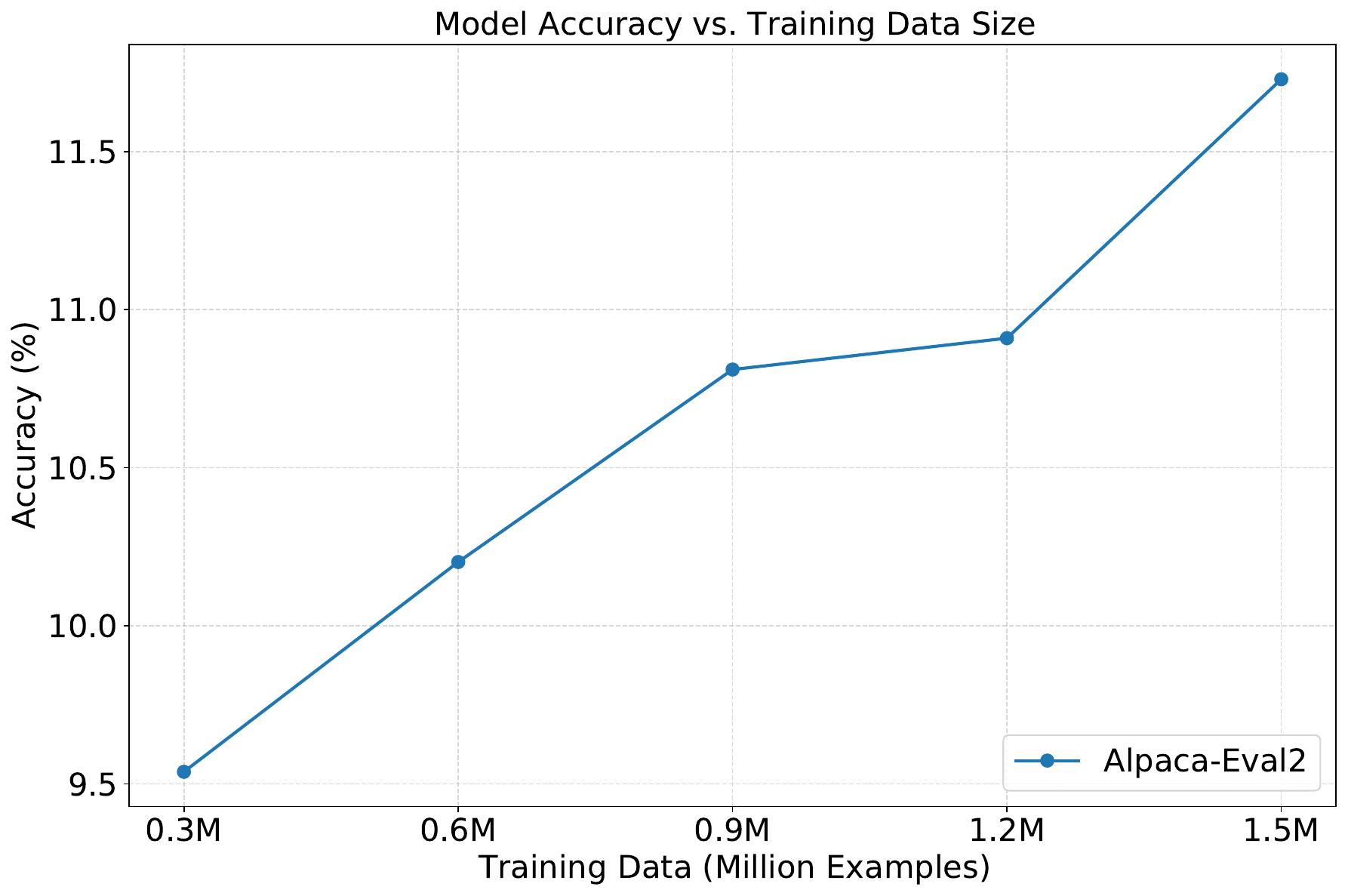}}
\caption {Scaling curves on foundational and conversational tasks.}
\label{scaling}
\end{figure*}

To further explore the correlation between the data quantity and model performance in foundational and conversational tasks, we control the full amount of data at different scales to fine-tune the Qwen2.5\_1.5B and observe the corresponding change in the model performance on downstream tasks. Unbalanced sampling would cause large fluctuations in experimental results. Therefore, we labeled all instruction responses with reward scores using the reward model. With the instruction labeling system provided in Section \ref{synthesis}, we sorted all instructions under each second-level labeling type based the labeled reward scores. During the sampling process, instructions with higher reward values under all second-level labels will be retained as much as possible. The effectiveness curves are shown in Figure~\ref{scaling}, and we can observe a continuous trend of increasing overall model performance with the number of instruction data in both foundational and conversational scenarios. Specifically, as the quantity of instruction data increases, there is a significant growth rate in the model's performance on reasoning tasks such as MATH and GSM-8K. This result illustrates that large-scale instruction datasets are necessary for building general-purpose language models. It also guides us in the direction of continuous optimization of the instruction dataset.  

\section{Ablation Study}
In the main experiments, we are using a two-stage training approach to maximize the effect of the foundational dataset and the conversational dataset. In this section, we will analyze the effect of training the foundational dataset or the conversational dataset alone and the benefits of two-stage training. Considering the training costs, we choose an earlier version of the foundational instruction dataset, InfInstruct-F-3M, and an earlier version of the conversational instruction dataset, InfInstruct-G-300K. We utilize mistral-7B as the base model. We evaluate the performance of each ablation setting based on the conversational performance of the finetuned model.

\textbf{Single foundational dataset or conversational dataset}. As shown in Table \ref{two stage}, we find that training on the dialog dataset alone yields 93.3\% of the performance of full two-stage training. Although training on the foundational dataset alone does not result in the highest level of conversational ability, it can further enhance the conversational ability of the language model in the two-stage training settings.

\textbf{Optimal trade-off of foundational and conversational ability}. Since the foundational dataset and the conversational dataset focus on the improvement of the foundational ability and the improvement of the conversational ability, respectively. We considered two ways to balance these two improvements. The first is that we break up and mix the foundational dataset and the conversational dataset. Then we perform one-stage training. The second way refers to the idea of curriculum learning~\cite{bengio2009curriculum}, where we train on the foundational dataset first, and then on the conversational instruction dataset. The results are shown in Table \ref{two stage}, we can see that simply merging datasets and doing an end-to-end training does not lead to higher conversational ability or foundational ability. Instead, two-stage training leads to the highest conversational ability and a higher foundational ability than one-stage training. This demonstrates that models with stronger foundational capabilities can also achieve stronger conversational capabilities.

\begin{table}[htbp]
\caption{InfInstruct-F-7.4M and InfInstruct-F-1.5M Dialog Turns Statistics}
\centering
\begin{tabular}{cc|c|c|c|c}
\hline
\multicolumn{2}{c|}{\textbf{Turns}}                              & \textbf{1}       & \textbf{(1, 5{]}} & \textbf{(5, 10{]}} & \textbf{(10, $\infty${)}}  \\ \hline
\multicolumn{1}{c|}{\multirow{2}{*}{F-7.4M}} & Samples    & 6897934 & 466354  & 73912    & 10906 \\ 
\multicolumn{1}{c|}{}                      & Percentage & 0.926   & 0.063   & 0.010    & 0.001 \\ \hline
\multicolumn{1}{c|}{\multirow{2}{*}{G-1.5M}} & Samples    & 1423191 & 28211   & 4647     & 878   \\ 
\multicolumn{1}{c|}{}                      & Percentage & 0.977   & 0.019   & 0.003    & 0.001 \\ \hline
\end{tabular}
\end{table}
\label{turns}

\begin{table}[tb]
\caption{Infinity Instruct compares the effectiveness of open-source datasets in improving the foundational capabilities of pre-trained models. The best scores are bolded.}
\small
\begin{tabular}{lccccccc}
\toprule
\multicolumn{1}{c}{\multirow{2}{*}{\textbf{Dataset}}} & \multicolumn{2}{c}{\textbf{Math}} & \multicolumn{2}{c}{\textbf{Code}}  & \multicolumn{2}{c}{\textbf{Knowledge}} & \multicolumn{1}{l}{\textbf{Overall}} \\ \cline{2-7}
\multicolumn{1}{c}{}                                  & \textbf{MATH}  & \textbf{GSM-8K}  & \textbf{HumanEval} & \textbf{MBPP} & \textbf{MMLU}     & \textbf{C-EVAL}    & \multicolumn{1}{l}{}                 \\ \hline
\multicolumn{7}{l}{\textbf{Official-Version}}                                                                                                                           & \multicolumn{1}{l}{}                 \\ 
\midrule
Llama3.1-8B                                           & 13.4           & 50.1             & 25.6               & 45.2          & 49.3              & 32.3               & 36.0                                 \\
Llama3.1-8B-Instruct                                  & 22.3           & 67.8             & 52.1               & 50.5          & 53.2              & 47.5               & 48.9                                 \\ 
\midrule
\multicolumn{7}{l}{\textbf{SFT with Open-source Dataset}}                                                                                                               & \multicolumn{1}{l}{}                 \\ 
\midrule
w/ WildChat                                           & 10.2           & 52.7             & 49.2               & 48.2          & 55.8              & 41.1               & 42.9                                 \\
w/ Lmsys-conversational-1m                                      & 12.5           & 29.4             & 43.1               & 43.7          & 50.2              & 50.4               & 38.2                                 \\
w/ Evol-Instruct                                      & 22.1           & 54.0             & 53.1               & 43.1          & 65.0              & 48.3               & 47.6                                 \\
w/ UltraChat                                          & 15.0           & 59.4             & 54.8               & 44.3          & 65.4              & 52.7               & 48.6                                 \\
w/ MAGPIE-Pro                                         & 21.6           & 68.7             & 58.2               & 50.3          & 65.1              & 49.0               & 52.1                                 \\
w/ OpenHermes-2.5                                     & 9.9            & 73.4             & 56.2               & 49.1          & 64.7              & 49.8               & 50.5                                 \\
\textbf{w/ Infinity-Instruct}                         & \textbf{26.0}  & \textbf{78.6}    & \textbf{59.8}      & \textbf{52.0} & \textbf{66.2}     & \textbf{56.3}      & \textbf{56.5}                        \\ 
\bottomrule
\end{tabular}
\label{main-result-2}
\end{table}

\begin{table}[h!]
\caption{Ablation study of two-stage training, the foundational dataset, and the conversational dataset. In the One-Stage setup, we break up and mix the InfInstruct-F-3M and the InfInstruct-G-300K. We tested the results of AlpacaEval 2.0 for conversational ability evaluation, tested all the benchmarks in Table \ref{main-result-2}, and calculated the mean values for foundational ability evaluation.}
\centering
\begin{tabular}{l|cc}
\hline
\textbf{Settings} & \textbf{Conversational Ability}             & \textbf{Foundational Ability}    \\ \hline
InfInstruct-F-3M     & 16.0  &  \textbf{54.4}                        \\
InfInstruct-G-300K     & 23.8  & 43.9 \\
One-Stage & 22.3 & 50.9 \\
Two-Stage & \textbf{25.5} & 52.0 \\ \hline
\end{tabular}

\label{two stage}
\end{table}


\section{Conclusion}
In this paper, we introduced Infinity Instruct, a comprehensive approach to advancing open-source instruction datasets and bridging the performance gap with proprietary models. By addressing the limitations of existing datasets, which often prioritize narrow task domains, our work emphasizes high-quality data curation and synthesis to support both foundational and conversational tasks.

The pipeline of Infinity Instruct demonstrates the importance of robust data selection strategies, as evidenced by InfInstruct-F-7M, a foundational dataset selected from over 100M samples, and InfInstruct-G-1.5M, a conversational dataset designed with a two-layer labeling system. Through rigorous quality assurance techniques such as de-duplication and contamination detection, we ensured the reliability of our datasets across diverse domains, including mathematics, coding, and knowledge-based Q\&A. We fine-tune multiple open-source models on Infinity Instruct and achieve state-of-the-art results, showcasing significant improvements in both foundational and conversational benchmarks. Notably, InfInstruct-Llama3.1-70B achieved remarkable performance, surpassing GPT-4-0314 by 8.6\% in chat ability while maintaining near parity in foundational tasks. These results underscore the transformative potential of open-source initiatives in fostering accessible and high-performing language models. We hope that Infinity Instruct serves as a valuable resource and catalyst for further innovation within the broader research community.

\clearpage
\bibliographystyle{reference}
\bibliography{reference}

\begin{thebibliography}{31}
\providecommand{\natexlab}[1]{#1}
\providecommand{\url}[1]{\texttt{#1}}
\expandafter\ifx\csname urlstyle\endcsname\relax
  \providecommand{\doi}[1]{doi: #1}\else
  \providecommand{\doi}{doi: \begingroup \urlstyle{rm}\Url}\fi

\bibitem[Austin et~al.(2021)Austin, Odena, Nye, Bosma, Michalewski, Dohan,
  Jiang, Cai, Terry, Le, et~al.]{austin2021program}
Jacob Austin, Augustus Odena, Maxwell Nye, Maarten Bosma, Henryk Michalewski,
  David Dohan, Ellen Jiang, Carrie Cai, Michael Terry, Quoc Le, et~al.
\newblock Program synthesis with large language models.
\newblock \emph{arXiv preprint arXiv:2108.07732}, 2021.

\bibitem[Bengio et~al.(2009)Bengio, Louradour, Collobert, and
  Weston]{bengio2009curriculum}
Yoshua Bengio, J{\'e}r{\^o}me Louradour, Ronan Collobert, and Jason Weston.
\newblock Curriculum learning.
\newblock In \emph{Proceedings of the 26th annual international conference on
  machine learning}, pp.\  41--48, 2009.

\bibitem[Brown(2020)]{brown2020language}
Tom~B Brown.
\newblock Language models are few-shot learners.
\newblock \emph{arXiv preprint arXiv:2005.14165}, 2020.

\bibitem[Chen et~al.(2021)Chen, Tworek, Jun, Yuan, Pinto, Kaplan, Edwards,
  Burda, Joseph, Brockman, et~al.]{chen2021evaluating}
Mark Chen, Jerry Tworek, Heewoo Jun, Qiming Yuan, Henrique Ponde De~Oliveira
  Pinto, Jared Kaplan, Harri Edwards, Yuri Burda, Nicholas Joseph, Greg
  Brockman, et~al.
\newblock Evaluating large language models trained on code.
\newblock \emph{arXiv preprint arXiv:2107.03374}, 2021.

\bibitem[Cobbe et~al.(2021)Cobbe, Kosaraju, Bavarian, Chen, Jun, Kaiser,
  Plappert, Tworek, Hilton, Nakano, et~al.]{cobbe2021training}
Karl Cobbe, Vineet Kosaraju, Mohammad Bavarian, Mark Chen, Heewoo Jun, Lukasz
  Kaiser, Matthias Plappert, Jerry Tworek, Jacob Hilton, Reiichiro Nakano,
  et~al.
\newblock Training verifiers to solve math word problems.
\newblock \emph{arXiv preprint arXiv:2110.14168}, 2021.

\bibitem[Contributors(2023)]{2023opencompass}
OpenCompass Contributors.
\newblock Opencompass: A universal evaluation platform for foundation models.
\newblock \url{https://github.com/open-compass/opencompass}, 2023.

\bibitem[Ding et~al.(2023)Ding, Chen, Xu, Qin, Zheng, Hu, Liu, Sun, and
  Zhou]{ding2023enhancing}
Ning Ding, Yulin Chen, Bokai Xu, Yujia Qin, Zhi Zheng, Shengding Hu, Zhiyuan
  Liu, Maosong Sun, and Bowen Zhou.
\newblock Enhancing chat language models by scaling high-quality instructional
  conversations.
\newblock \emph{arXiv preprint arXiv:2305.14233}, 2023.

\bibitem[Hendrycks et~al.(2020)Hendrycks, Burns, Basart, Zou, Mazeika, Song,
  and Steinhardt]{hendrycks2020measuring}
Dan Hendrycks, Collin Burns, Steven Basart, Andy Zou, Mantas Mazeika, Dawn
  Song, and Jacob Steinhardt.
\newblock Measuring massive multitask language understanding.
\newblock \emph{arXiv preprint arXiv:2009.03300}, 2020.

\bibitem[Hendrycks et~al.(2021)Hendrycks, Burns, Kadavath, Arora, Basart, Tang,
  Song, and Steinhardt]{hendrycks2021measuring}
Dan Hendrycks, Collin Burns, Saurav Kadavath, Akul Arora, Steven Basart, Eric
  Tang, Dawn Song, and Jacob Steinhardt.
\newblock Measuring mathematical problem solving with the math dataset.
\newblock \emph{arXiv preprint arXiv:2103.03874}, 2021.

\bibitem[Huang et~al.(2023)Huang, Bai, Zhu, Zhang, Zhang, Su, Liu, Lv, Zhang,
  Lei, Fu, Sun, and He]{huang2023ceval}
Yuzhen Huang, Yuzhuo Bai, Zhihao Zhu, Junlei Zhang, Jinghan Zhang, Tangjun Su,
  Junteng Liu, Chuancheng Lv, Yikai Zhang, Jiayi Lei, Yao Fu, Maosong Sun, and
  Junxian He.
\newblock C-eval: A multi-level multi-discipline chinese evaluation suite for
  foundation models.
\newblock In \emph{Advances in Neural Information Processing Systems}, 2023.

\bibitem[Li et~al.(2023{\natexlab{a}})Li, Zhang, Li, Chen, Chen, Cheng, Wang,
  Zhou, and Xiao]{li2023quantity}
Ming Li, Yong Zhang, Zhitao Li, Jiuhai Chen, Lichang Chen, Ning Cheng, Jianzong
  Wang, Tianyi Zhou, and Jing Xiao.
\newblock From quantity to quality: Boosting llm performance with self-guided
  data selection for instruction tuning.
\newblock \emph{arXiv preprint arXiv:2308.12032}, 2023{\natexlab{a}}.

\bibitem[Li et~al.(2024{\natexlab{a}})Li, Zhang, He, Li, Zhao, Wang, Cheng, and
  Zhou]{li2024superfiltering}
Ming Li, Yong Zhang, Shwai He, Zhitao Li, Hongyu Zhao, Jianzong Wang, Ning
  Cheng, and Tianyi Zhou.
\newblock Superfiltering: Weak-to-strong data filtering for fast
  instruction-tuning.
\newblock \emph{arXiv preprint arXiv:2402.00530}, 2024{\natexlab{a}}.

\bibitem[Li et~al.(2024{\natexlab{b}})Li, Chiang, Frick, Dunlap, Zhu,
  et~al.]{arenahard2024}
Tianle Li, Wei-Lin Chiang, Evan Frick, Lisa Dunlap, Banghua Zhu, et~al.
\newblock From live data to high-quality benchmarks: The arena-hard pipeline,
  April 2024{\natexlab{b}}.
\newblock URL \url{https://lmsys.org/blog/2024-04-19-arena-hard/}.

\bibitem[Li et~al.(2023{\natexlab{b}})Li, Zhang, Dubois, Taori, Gulrajani,
  Guestrin, Liang, and Hashimoto]{alpaca_eval}
Xuechen Li, Tianyi Zhang, Yann Dubois, Rohan Taori, Ishaan Gulrajani, Carlos
  Guestrin, Percy Liang, and Tatsunori~B. Hashimoto.
\newblock Alpacaeval: An automatic evaluator of instruction-following models.
\newblock \url{https://github.com/tatsu-lab/alpaca_eval}, 5 2023{\natexlab{b}}.

\bibitem[Longpre et~al.(2023)Longpre, Hou, Vu, Webson, Chung, Tay, Zhou, Le,
  Zoph, Wei, et~al.]{longpre2023flan}
Shayne Longpre, Le~Hou, Tu~Vu, Albert Webson, Hyung~Won Chung, Yi~Tay, Denny
  Zhou, Quoc~V Le, Barret Zoph, Jason Wei, et~al.
\newblock The flan collection: Designing data and methods for effective
  instruction tuning.
\newblock \emph{arXiv preprint arXiv:2301.13688}, 2023.

\bibitem[Lu et~al.(2023)Lu, Yuan, Yuan, Lin, Lin, Tan, Zhou, and
  Zhou]{lu2023instag}
Keming Lu, Hongyi Yuan, Zheng Yuan, Runji Lin, Junyang Lin, Chuanqi Tan, Chang
  Zhou, and Jingren Zhou.
\newblock \# instag: Instruction tagging for analyzing supervised fine-tuning
  of large language models.
\newblock In \emph{The Twelfth International Conference on Learning
  Representations}, 2023.

\bibitem[Ouyang et~al.(2022)Ouyang, Wu, Jiang, Almeida, Wainwright, Mishkin,
  Zhang, Agarwal, Slama, Ray, et~al.]{ouyang2022training}
Long Ouyang, Jeffrey Wu, Xu~Jiang, Diogo Almeida, Carroll Wainwright, Pamela
  Mishkin, Chong Zhang, Sandhini Agarwal, Katarina Slama, Alex Ray, et~al.
\newblock Training language models to follow instructions with human feedback.
\newblock \emph{Advances in neural information processing systems},
  35:\penalty0 27730--27744, 2022.

\bibitem[Radford et~al.(2019)Radford, Wu, Child, Luan, Amodei, Sutskever,
  et~al.]{radford2019language}
Alec Radford, Jeffrey Wu, Rewon Child, David Luan, Dario Amodei, Ilya
  Sutskever, et~al.
\newblock Language models are unsupervised multitask learners.
\newblock \emph{OpenAI blog}, 1\penalty0 (8):\penalty0 9, 2019.

\bibitem[Teknium(2023)]{OpenHermes25}
Teknium.
\newblock Openhermes 2.5: An open dataset of synthetic data for generalist llm
  assistants, 2023.
\newblock URL \url{https://huggingface.co/datasets/teknium/OpenHermes-2.5}.

\bibitem[Xiao et~al.(2023)Xiao, Liu, Zhang, and Muennighoff]{bge_embedding}
Shitao Xiao, Zheng Liu, Peitian Zhang, and Niklas Muennighoff.
\newblock C-pack: Packaged resources to advance general chinese embedding,
  2023.

\bibitem[Xie et~al.(2023)Xie, Santurkar, Ma, and
  Liang]{xie2023dataselectionlanguagemodels}
Sang~Michael Xie, Shibani Santurkar, Tengyu Ma, and Percy Liang.
\newblock Data selection for language models via importance resampling, 2023.
\newblock URL \url{https://arxiv.org/abs/2302.03169}.

\bibitem[Xu et~al.(2023)Xu, Sun, Zheng, Geng, Zhao, Feng, Tao, and
  Jiang]{xu2023wizardlm}
Can Xu, Qingfeng Sun, Kai Zheng, Xiubo Geng, Pu~Zhao, Jiazhan Feng, Chongyang
  Tao, and Daxin Jiang.
\newblock Wizardlm: Empowering large language models to follow complex
  instructions.
\newblock \emph{arXiv preprint arXiv:2304.12244}, 2023.

\bibitem[Xu et~al.(2024)Xu, Jiang, Niu, Deng, Poovendran, Choi, and
  Lin]{xu2024magpie}
Zhangchen Xu, Fengqing Jiang, Luyao Niu, Yuntian Deng, Radha Poovendran, Yejin
  Choi, and Bill~Yuchen Lin.
\newblock Magpie: Alignment data synthesis from scratch by prompting aligned
  llms with nothing.
\newblock \emph{arXiv preprint arXiv:2406.08464}, 2024.

\bibitem[Yang et~al.(2024)Yang, Yang, Hui, Zheng, Yu, Zhou, Li, Li, Liu, Huang,
  et~al.]{yang2024qwen2}
An~Yang, Baosong Yang, Binyuan Hui, Bo~Zheng, Bowen Yu, Chang Zhou, Chengpeng
  Li, Chengyuan Li, Dayiheng Liu, Fei Huang, et~al.
\newblock Qwen2 technical report.
\newblock \emph{arXiv preprint arXiv:2407.10671}, 2024.

\bibitem[Zhang et~al.(2024{\natexlab{a}})Zhang, Wang, Li, Gu, Wu, Zhang, Gao,
  Ao, and Liu]{zhang2024aquila2}
Bo-Wen Zhang, Liangdong Wang, Jijie Li, Shuhao Gu, Xinya Wu, Zhengduo Zhang,
  Boyan Gao, Yulong Ao, and Guang Liu.
\newblock Aquila2 technical report.
\newblock \emph{arXiv preprint arXiv:2408.07410}, 2024{\natexlab{a}}.

\bibitem[Zhang et~al.(2024{\natexlab{b}})Zhang, Yan, Li, and
  Liu]{zhang2024infinitymath}
Bo-Wen Zhang, Yan Yan, Lin Li, and Guang Liu.
\newblock Infinitymath: A scalable instruction tuning dataset in programmatic
  mathematical reasoning.
\newblock In \emph{Proceedings of the 33rd ACM International Conference on
  Information and Knowledge Management}, pp.\  5405--5409, 2024{\natexlab{b}}.

\bibitem[Zhang et~al.(2024{\natexlab{c}})Zhang, Qu, Liu, Zhang, Lin, Yu, Pan,
  Cheng, Liu, Lin, et~al.]{zhang2024map}
Ge~Zhang, Scott Qu, Jiaheng Liu, Chenchen Zhang, Chenghua Lin, Chou~Leuang Yu,
  Danny Pan, Esther Cheng, Jie Liu, Qunshu Lin, et~al.
\newblock Map-neo: Highly capable and transparent bilingual large language
  model series.
\newblock \emph{arXiv preprint arXiv:2405.19327}, 2024{\natexlab{c}}.

\bibitem[Zhao et~al.(2024{\natexlab{a}})Zhao, Du, Ju, Wu, and
  Pan]{zhao2024iidoptimizinginstructionlearning}
Hanyu Zhao, Li~Du, Yiming Ju, Chengwei Wu, and Tengfei Pan.
\newblock Beyond iid: Optimizing instruction learning from the perspective of
  instruction interaction and dependency.
\newblock 2024{\natexlab{a}}.
\newblock URL \url{https://arxiv.org/abs/2409.07045}.

\bibitem[Zhao et~al.(2024{\natexlab{b}})Zhao, Ren, Hessel, Cardie, Choi, and
  Deng]{zhao2024wildchat}
Wenting Zhao, Xiang Ren, Jack Hessel, Claire Cardie, Yejin Choi, and Yuntian
  Deng.
\newblock Wildchat: 1m chatgpt interaction logs in the wild.
\newblock \emph{arXiv preprint arXiv:2405.01470}, 2024{\natexlab{b}}.

\bibitem[Zheng et~al.(2023{\natexlab{a}})Zheng, Chiang, Sheng, Li, Zhuang, Wu,
  Zhuang, Li, Lin, Xing, Gonzalez, Stoica, and Zhang]{zheng2023lmsyschat1m}
Lianmin Zheng, Wei-Lin Chiang, Ying Sheng, Tianle Li, Siyuan Zhuang, Zhanghao
  Wu, Yonghao Zhuang, Zhuohan Li, Zi~Lin, Eric.~P Xing, Joseph~E. Gonzalez, Ion
  Stoica, and Hao Zhang.
\newblock Lmsys-chat-1m: A large-scale real-world llm conversation dataset,
  2023{\natexlab{a}}.

\bibitem[Zheng et~al.(2023{\natexlab{b}})Zheng, Chiang, Sheng, Zhuang, Wu,
  Zhuang, Lin, Li, Li, Xing, et~al.]{zheng2023judging}
Lianmin Zheng, Wei-Lin Chiang, Ying Sheng, Siyuan Zhuang, Zhanghao Wu, Yonghao
  Zhuang, Zi~Lin, Zhuohan Li, Dacheng Li, Eric Xing, et~al.
\newblock Judging llm-as-a-judge with mt-bench and chatbot arena.
\newblock \emph{Advances in Neural Information Processing Systems},
  36:\penalty0 46595--46623, 2023{\natexlab{b}}.

\end{thebibliography}


\clearpage
\appendix
\section{Evaluation Setup}
\textbf{Chat ability evaluation}. 
We evaluate the instruction-following capabilities of fine-tuned models on three leading benchmarks:
\begin{itemize}
    \item Mt-Bench~\cite{zheng2023judging}: A set of multi-turn questions across 8 categories, including writing, role-play, extraction, reasoning, math, coding, knowledge (STEM), and knowledge (humanities/social science). It uses GPT-4 as a referee and baseline, evaluating the model's response win rate relative to GPT-4.
    \item AlpacaEval2.0~\cite{alpaca_eval}: Comprising 805 prompts from the AlpacaFarm evaluation set, this benchmark uses GPT-4 as a referee and baseline. It introduces length-controlled win rates to reduce GPT-4's length bias and aligns closely with the human-annotated Chatbot Arena.
    \item Arena-Hard~\cite{arenahard2024}: An automatic evaluation tool with 500 challenging user queries, using GPT-4-Turbo as a judge to compare models' responses to GPT-4. Arena-Hard exhibits the highest correlation and separability with Chatbot Arena among popular open-ended LLM benchmarks.
\end{itemize}

\textbf{Foundational ability evaluation}.
We evaluated the fine-tuned model across several datasets using the OpenCompass framework~\cite{2023opencompass}:
\begin{itemize}
    \item MMLU~\cite{hendrycks2020measuring}: A large-scale benchmark covering diverse tasks (history, math, science) to assess reasoning and knowledge.
    \item C-EVAL~\cite{huang2023ceval}: A Chinese language dataset for tasks like question answering, summarization, and translation.
    \item MATH~\cite{hendrycks2021measuring}: A dataset with high school-level math problems requiring multi-step reasoning.
    \item GSM8K~\cite{cobbe2021training}: A dataset of elementary math problems to evaluate basic arithmetic and logical reasoning.
    \item HumanEval~\cite{chen2021evaluating}: A Python programming dataset for testing code generation models.
    \item MBPP~\cite{austin2021program}: A dataset for assessing basic Python programming tasks, including task descriptions, code, and tests.
\end{itemize}
These datasets comprehensively evaluate the model’s reasoning, mathematical, and coding abilities.

\section{Related works}
The advent of Large Pre-trained Language Models (LLMs) has marked a significant shift in the field of Natural Language Processing (NLP). These models, such as GPT-3 and its successors, have demonstrated remarkable capabilities in understanding and generating human-like text across a variety of tasks~\cite{brown2020language}. However, the performance of these models is heavily influenced by the quality and scale of the data they are fine-tuned on ~\cite{radford2019language}.

The importance of high-quality instruction datasets for fine-tuning LLMs has been underscored by several studies. For instance, the work on Superfiltering by ~\cite{li2024superfiltering} highlights the need for weak-to-strong data filtering to expedite instruction-tuning. Similarly, \cite{xu2023wizardlm} introduced WizardLM, which focuses on empowering LLMs to follow complex instructions by enhancing their understanding through a carefully curated dataset.

In the open-source community, there have been efforts to compile datasets that could be used for fine-tuning purposes. Projects like OpenHermes-2.5 and CodeBagel have contributed to this space by providing datasets that encompass a wide range of instructions \cite{OpenHermes25}. However, the challenge lies in the curation of these datasets to ensure they are not only extensive but also maintain high standards of quality and relevance \cite{lu2023instag}.

Infinity Instruct addresses this gap by introducing a large-scale, high-quality instruction dataset. This project stands out due to its focus on both the foundational dataset, which contains millions of instructions selected from various open-source datasets, and the chat dataset, evolved from a subset of high-quality seed data to improve real conversation scenarios.

Moreover, the project's approach to data evolution and model ability diagnosis represents a novel contribution to the field. By systematically tagging instructions and using AI assistants for data validation and generation, "Infinity Instruct" ensures that the dataset is not only large but also diverse and informative, catering to the complex needs of LLMs \cite{zhang2024map}.

\section{Training Details}
The training process consists of two phases. We first apply the foundational dataset Infinity-Instruct-7M to improve the foundational ability (e.g. math, code) of the pre-trained models. We further fine-tune the dialogue dataset to improve the model's chat ability. We found that the training hyperparameters are more relevant to some attributes of the pre-trained model, such as the number of parameters, the end learning rate, etc. Thus, we apply the same training hyperparameters to both phases of training. The main training hyperparameters are shown in Table \ref{Hyperparameters}. To improve the training efficiency on large-scale instruction datasets, we apply FlagScale, a distributed training framework. This framework compresses the number of training steps through the packing strategy. Meanwhile, it applies multiple model-parallel and data-parallel techniques to accelerate the training of tens of billions of models.
\begin{table}[tb]
\centering
\small
\caption{Training hyperparameters for all models in both phases.}
\begin{tabular}{lccccc}
\toprule
\textbf{Model}  & \textbf{Epoch} & \textbf{LR} & \multicolumn{1}{l}{\textbf{Context Length}} & \textbf{Warmup Steps} & \textbf{Batch Size} \\ 
\midrule
Llama-3.1-8B     & 3              & 5e-6        & 4096                                        & 40                    & 528
\\ 
Llama-3.1-70B     & 3              & 5e-6        & 4096                                        & 40                    & 528                 \\ 
Mistral-7B-v0.1 & 3              & 5e-6        & 4096                                        & 40                    & 528                 \\
Qwen-2-7B       & 3              & 1e-5        & 4096                                        & 40                    & 528                 \\ 
Yi-1.5-9B       & 3              & 1e-5        & 4096                                        & 40                    & 528                 \\ 
\bottomrule
\end{tabular}
\label{Hyperparameters}
\end{table}

\begin{figure}[t]
    \centering
    \includegraphics[width=0.9\linewidth]{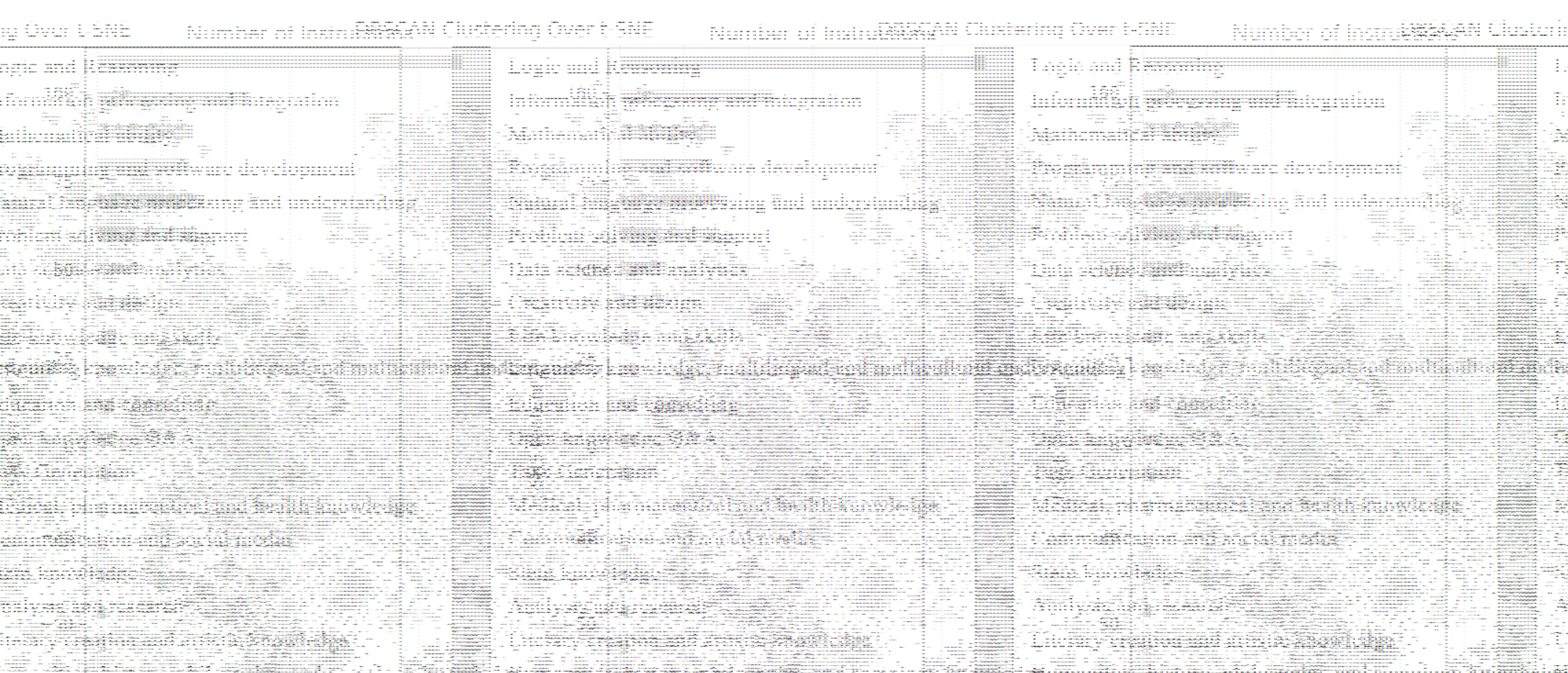}
    \caption{T-SNE visualization and analysis of the first-level label type distribution of instructions in the InfInstruct-F-7.4M and InfInstruct-G-1.5M dataset. We sampled up to 2000 instructions per label type.}
    \label{fig:data-cluster}
\label{tsne}
\end{figure}

\section{Data Analysis}
We conducted a statistical analysis of InfInstruct-F-7.4M and InfInstruct-G-1.5M, with results presented in Figure \ref{tsne}. To visualize the semantic distribution of instructions, we applied the t-SNE algorithm to both datasets. The resulting distribution graphs reveal that certain challenging and complex task types, such as Logic and Reasoning, exhibit a more dispersed distribution. These tasks often involve a broader range of knowledge points and demand intricate problem-solving capabilities. As discussed in Section \ref{data-select}, we posit that enhancing the diversity of instructions can significantly improve the model's baseline performance on such complex tasks. To this end, InfInstruct-F-7.4M incorporates a substantial number of instructions categorized under "Information Processing". Conversely, InfInstruct-G-1.5M emphasizes a balanced allocation of instructions across different task types, ensuring the model demonstrates robust instruction-following capabilities across a variety of downstream applications.

Additionally, we analyzed the dialogue turn statistics for InfInstruct-F-7.4M and InfInstruct-G-1.5M, summarized in Table \ref{turns}. Single-turn dialogues accounted for 92.6\% and 97.7\% of the instructions in InfInstruct-F-7.4M and InfInstruct-G-1.5M, respectively. To enhance the model's adaptability to multi-turn dialogue scenarios, we supplemented the datasets with approximately 7.4\% and 2.3\% multi-turn instructions, respectively. This strategic augmentation supports the development of more effective multi-turn dialogue capabilities in downstream tasks.

\section{Limitation}
Limitations. While Infinity-Instruct significantly improves the performance of open-source LLMs, several limitations remain. First, the labeling and selection process relies on existing open-source models, which may introduce bias or miss nuanced instruction types not well-represented in the original data pool. Second, although our pipeline is scalable, it still requires substantial computational resources for clustering, synthesis, and fine-tuning. Lastly, our evaluation focuses primarily on benchmark performance; real-world robustness, safety, and long-term retention of capabilities across domains warrant further investigation.

\section{Broader Impacts}
Infinity-Instruct lowers the barrier to building high-quality instruction-tuned models by providing an open, scalable framework for dataset construction. This helps democratize access to capable LLMs, especially for researchers and organizations with limited resources. However, as the pipeline relies on existing data and models, it may inherit biases or limitations from them. Careful dataset auditing and responsible model deployment are essential to ensure ethical and safe use.


\clearpage

\end{document}